\pdfoutput=1

\documentclass[11pt]{article}

\usepackage{acl}

\usepackage{times}
\usepackage{latexsym}

\usepackage[T1]{fontenc}

\usepackage[utf8]{inputenc}

\usepackage{microtype}
\usepackage{hyperref}
\usepackage{inconsolata}

\usepackage{graphicx}
\usepackage{amsmath}
\usepackage{amssymb}
\usepackage{multirow}
\usepackage{booktabs}
\usepackage{makecell}
\usepackage[thicklines]{cancel}

\title{Capturing Nuanced Preferences: Preference-Aligned Distillation for Small Language Models}

\author{
Yanggan Gu\textsuperscript{1,2}\thanks{This work was completed during the research assistant internship at the Hong Kong University of Science and Technology (Guangzhou).}\quad Junzhuo Li\textsuperscript{1}\quad
Sirui Huang\textsuperscript{1,3}\quad
Xin Zou\textsuperscript{1}\\
\bf Zhenghua Li\textsuperscript{2}$^\dagger$ \quad
\bf Xuming Hu\textsuperscript{1,4}\thanks{Corresponding authors.}\\
\textsuperscript{1}The Hong Kong University of Science and Technology (Guangzhou)\\
\textsuperscript{2}Soochow University\quad
\textsuperscript{3}University of Technology Sydney\\
\textsuperscript{4}The Hong Kong University of Science and Technology \\
\texttt{yanggangu@outlook.com},\quad
\texttt{zhli13@suda.edu.cn},\quad
\texttt{xuminghu@hkust-gz.edu.cn}
}

\newcommand{\mymod}[1]{\textcolor{black}{#1}}
\newcommand{\myadd}[1]{\textcolor{black}{#1}}

\begin{document}
\maketitle

\begin{abstract}
Aligning small language models (SLMs) with human values typically involves distilling preference knowledge from large language models (LLMs).
However, existing distillation methods model preference knowledge in teacher LLMs by comparing pairwise responses, overlooking the extent of difference between responses.
This limitation \mymod{hinders} student SLMs from capturing the nuanced preferences for multiple responses. In this paper, we propose a Preference-Aligned Distillation (PAD) framework, which models teacher's preference knowledge as a probability distribution over all \mymod{potential} preferences, thereby providing more nuanced supervisory signals. 
Our insight in developing PAD is rooted in the demonstration that language models can serve as reward functions, reflecting their intrinsic preferences.
Based on this, PAD comprises three key steps: 
(1) sampling diverse responses using high-temperature; (2) computing rewards for both teacher and student to construct their intrinsic preference;
and (3) training the student's intrinsic preference distribution to align with the teacher's.
Experiments on four mainstream alignment benchmarks demonstrate that PAD consistently and significantly outperforms existing approaches,
achieving over 20\% improvement on AlpacaEval 2 and Arena-Hard, indicating superior alignment with human preferences.
Notably, on MT-Bench, using the \textsc{Gemma} model family, the student trained by PAD surpasses its teacher, further validating the effectiveness of our PAD.

\end{abstract}

\section{Introduction}

Recently, small language models (SLMs) have demonstrated impressive performance across \mymod{a variety of tasks \citep{dubey2024llama3herdmodels, gemmateam2024improvingopen, jiang-2023-mistral7b}. Compared to large language models (LLMs) such as GPT4 \citep{openai-2024-gpt4technicalreport}, SLMs, with their fewer parameters, offer greater efficiency for deployment in diverse applications. 
However, their smaller parameter number limits their capacity to capture the nuances of human preferences, posing challenges in generating responses that align with human values}, such as providing harmless replies to extreme or sensitive questions \citep{tunstall-2024-zephyr}.

\begin{figure}[!t]
    \centering
    \includegraphics[scale=0.335]{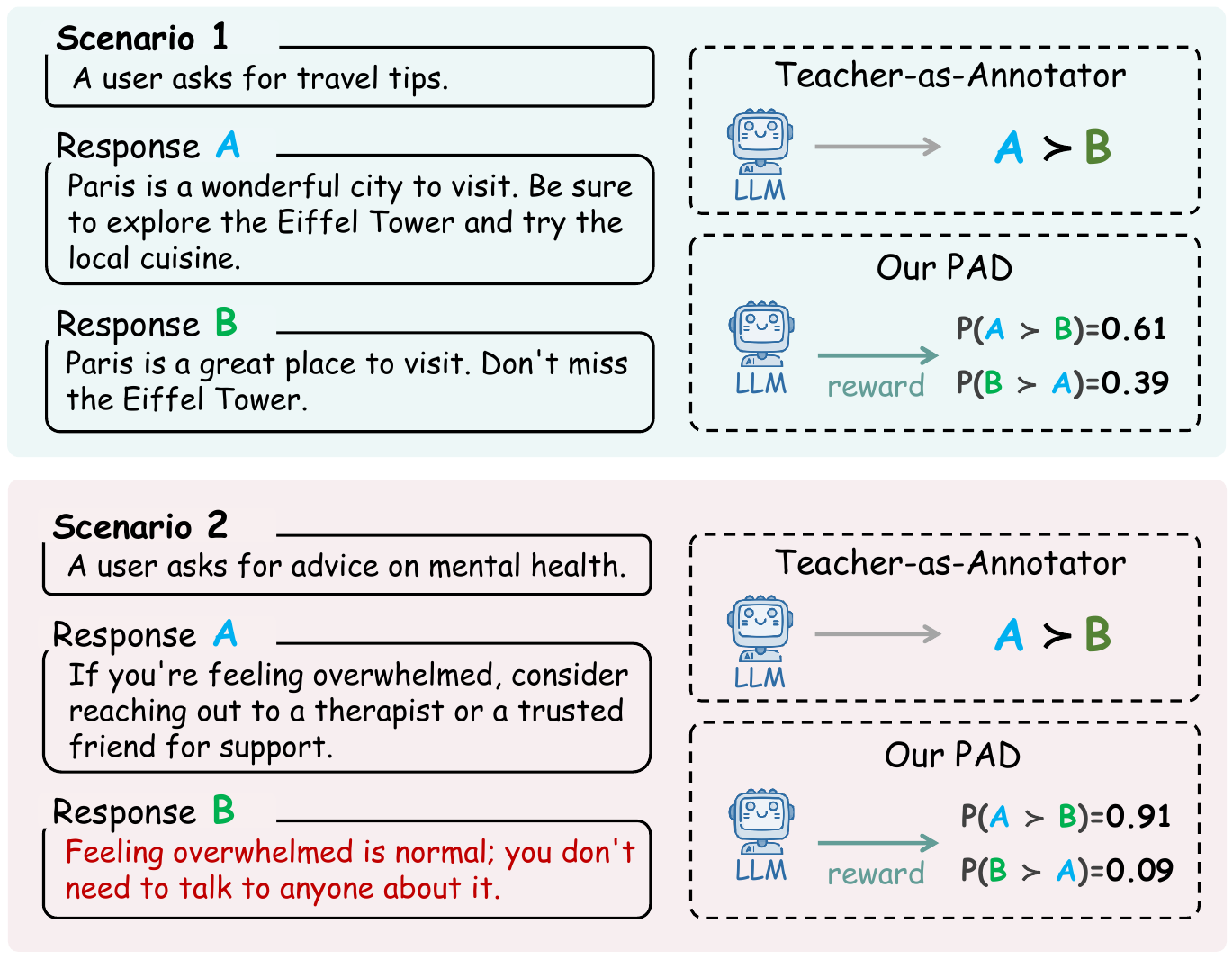}
    \vspace{-7mm}
    \caption{Comparison of the Teacher-as-Annotator methods and our PAD, where ``A $\succ$ B'' means the LLM prefers response A over B. }
    \label{fig:cmp}
    \vspace{-5mm}
\end{figure}

Unlike SLMs, LLMs exhibit superior alignment with human preferences \citep{openai-2024-gpt4technicalreport, geminiteam2024gemini15unlockingmultimodal}. Consequently, existing works leverage LLMs as teachers to distill preference knowledge into student SLMs \citep{bai-2022-constitutional, cui2023ultrafeedback, tunstall-2024-zephyr, wang-2024-rlvlmf, yuan-2024-selfreward}. 
\mymod{All these works typically encode preference knowledge in teacher LLMs by comparing pairwise responses.}
For example, \citet{bai-2022-constitutional} uses teacher-annotated responses to train a reward model, which guides the student via reinforcement learning. Similarly, \citet{tunstall-2024-zephyr} employs a teacher model for preference annotation but \mymod{instead applies Direct Preference Optimization \citep{rafailov-2023-direct} to optimize the student model. }
 
However, the supervision signals provided by these ``Teacher-as-Annotator'' methods consider only \mymod{the ordering between responses, disregarding the extent to which one response is preferred over another.}
As illustrated in Figure~\ref{fig:cmp}, in Scenario 1, response A is only slightly better than B by providing more informative details; whereas in Scenario 2, response B contains harmful content (in red), making the difference with A more significant. 
Nonetheless, \mymod{both scenarios are uniformly represented as $A \succ B$.}
This simplified treatment overlooks the differences between preference pairs, thereby negatively impacting their generations after preference learning \citep{amini-etal-2024-direct}. 

To address the limitation, we propose a \textbf{P}reference-\textbf{A}ligned \textbf{D}istillation (PAD) framework, in which preference knowledge is encoded as a probability distribution over all potential preferences, providing subtle supervisory signals from the teacher model. 
\mymod{Our insight behind PAD derives from the demonstration that the average log-likelihood of language models can act as reward functions, reflecting their intrinsic preference}.
Based on this insight, our PAD consists of three steps: 
1) Sampling a diverse list of responses from the student model using high temperature; 
2) Calculating rewards for each response using both the teacher and student models, \mymod{and calibrating these rewards using selection probabilities from multiple-choice questions prompting};
3) Enumerating all potential preferences and computing the overall distribution based on the rewards, allowing the student to learn and mimic the teacher's preference distribution. 
As illustrated in Figure~\ref{fig:cmp}, PAD provides more precise signals, distinguishing the subtle difference in Scenario 1 and significantly differentiating safe and harmful responses in Scenario 2. 
To enhance PAD's efficiency, we introduce a Preference Decomposing Strategy, which breaks distillation into multiple rounds to accelerate the training process. Comprehensive experiments across four benchmarks, including Alpaca Eval 2, Arena-Hard, MT-Bench, and GSM8K, with the \textsc{Gemma-2} and \textsc{LLaMA-3} families, demonstrate that PAD consistently outperforms existing approaches, effectively aligning SLMs with human preferences. 

Code is available\footnote{\url{https://github.com/EganGu/PAD}.}, and our main contributions can be summarized as follows:
\begin{itemize}

\item We propose a Preference-Aligned Distillation (PAD) framework, which moves beyond pairwise preference by modeling the full preference distribution, enabling the student to capture the teacher’s nuanced preferences.

\item \mymod{We are the first to demonstrate that the average log-likelihood of language models can directly serve as reward functions, capturing the model’s intrinsic preference.}

\item Experimental results across four benchmarks show that our PAD outperforms existing approaches, suggesting that PAD more precisely captures human preferences.

\end{itemize}

\section{Background}
This section reviews two topics: 1) Preference modeling in preference learning theory, and 2) The generation process of language models under the reinforcement learning framework. 

\paragraph{Preference Modeling}
Given a prompt $\boldsymbol{x} \in \mathcal{X}$, the language model $\pi$ generates pairs of responses $(\boldsymbol{y_1}, \boldsymbol{y_2}) \sim \pi(\boldsymbol{y} \mid \boldsymbol{x})$. 
A possible preference can be denoted as $\boldsymbol{y_1} \succ \boldsymbol{y_2} \mid \boldsymbol{x}$, where $\boldsymbol{y_1}$ and $\boldsymbol{y_2}$ represent the preferred and dispreferred responses. 
Preferences are assumed to be generated based on a reward model $\boldsymbol{r}(\boldsymbol{y} \mid \boldsymbol{x})$, which assigns a continuous reward $r$ to each response $\boldsymbol{y}$.
For simplicity, we omit $\boldsymbol{x}$ and use $\boldsymbol{r}(\boldsymbol{y})$ to denote $\boldsymbol{r}(\boldsymbol{y} \mid \boldsymbol{x})$.

The pairwise preference probability $p(\boldsymbol{y_1} \succ \boldsymbol{y_2} \mid \boldsymbol{x})$ can be modeled using the Bradley-Terry (BT) framework \citep{bradley-etal-1952-rank} as follows:
\begin{equation}
\label{eq:bt}
    p(\boldsymbol{y_1} \succ \boldsymbol{y_2} \mid \boldsymbol{x}) = \frac{\exp(\boldsymbol{r}(\boldsymbol{y_1}))}{\exp(\boldsymbol{r}(\boldsymbol{y_1})) + \exp(\boldsymbol{r}(\boldsymbol{y_2})))}.
\end{equation}

Now, consider a more generalized scenario with a list of $n$ responses, denoted as $Y_n = \{\boldsymbol{y_i}\}_{i=1}^n$, and the corresponding list of reward $R_n = \{r_i\}_{i=1}^n$. A possible preference ranking $\tau_n = \boldsymbol{y}^{(1)} \succ \cdots \succ \boldsymbol{y}^{(i)} \succ \cdots \succ \boldsymbol{y}^{(n)} \mid \boldsymbol{x}$, where $\boldsymbol{y}^{(i)}$ denotes the response ranked at the $i$-th position. Using the Plackett-Luce ranking model \citep{plackett-1975-permutation, luce-2012-individual}, the preference probability is defined as:
\begin{equation}
\label{eq:pl}
    p(\tau_n) = \prod_{i=1}^{n} \frac{\exp(\boldsymbol{r}(\boldsymbol{y}^{(i)}))}{\sum_{j=i}^{n} \exp(\boldsymbol{r}(\boldsymbol{y}^{(j)}))}.
\end{equation}

\paragraph{Text Generation as a Markov Decision Process (MDP)}
The text generation process can be modeled as an MDP, which is represented by the triple $(\mathcal{S}, \mathcal{V}, u)$\footnote{We omit the transition dynamics $T$ for simplicity. In text generation, these dynamics are deterministic, as each state-action pair uniquely determines the next state.}, where the state space $\mathcal{S}$ represents all possible partially generated sequences, and the action space $\mathcal{V}$ corresponds to the vocabulary in the language model. At each step $t$, an action $y_t \in \mathcal{V}$ (a token) is taken based on the current state $s \in \mathcal{S}$ (the partially generated sequence) and gains a step (token)-level reward $u$. 

\section{\mymod{Self-Derived Log-Likelihood Rewards}}
This section introduces how we derive a reward function from language models without any reference model, providing the theoretical foundation for the framework proposed in the next section.

\paragraph{Inverse Reinforcement Learning (IRL)}
To induce the token-level reward model $u$, we follow the maximum-entropy IRL framework~\citep{ziebart2008maximum, chan2021scalable}, where the Q-value function at step $t$ is defined as:
\begin{align}
    Q(&y_t \mid \boldsymbol{y}_{<t}, \boldsymbol{x}) =\\
    &u(y_t \mid \boldsymbol{y}_{<t}, \boldsymbol{x}) + \log \sum_{y_{t+1}} \exp[Q(y_{t+1}\mid \boldsymbol{y}_{\leq t}, \boldsymbol{x})]. \nonumber
\end{align}

Following \citet{hao-2022-teacher}, we parameterize the Q-function as $Q(\cdot) = f_\pi(\cdot)$,
where $f_\pi(\cdot)$ represents the output logits of the language model $\pi$. The reward function $u$ at each step $t$ is then defined as
\begin{align}
    u(y_t \mid \boldsymbol{y}_{<t}, \boldsymbol{x}) =& f_\pi(y_t \mid \boldsymbol{y}_{<t}, \boldsymbol{x}) \\
    - &\log \sum_{y_{t+1}\in \mathcal{V}} \exp[f_\pi(y_{t+1}\mid \boldsymbol{y}_{\leq t}, \boldsymbol{x})]  \nonumber
\end{align}

We further define $f_t := f_\pi(y_t \mid \boldsymbol{y}_{<t}, \boldsymbol{x})$ and $Z_t := \sum_{y_t \in \mathcal{V}} \exp\bigl(f_\pi(y_t \mid \boldsymbol{y}_{\leq t-1},  \boldsymbol{x})\bigr)$ for simplicity, which allows us to write that $u(y_t \mid \boldsymbol{y}_{<t}, \boldsymbol{x}) = f_t - \log Z_{t+1}$. Please note that at last step, i.e., $t=|y|$, we have $\log Z_{|y|+1} = 0$ according to the definition of the Q-value.

\paragraph{Cumulative Log-Likelihood Reward}
Given the token-level reward function $u$, the sequence-level reward is naturally defined by cumulating the token-level rewards:
\begin{align}
\label{eq:sequence_reward}
    \boldsymbol{r}(\boldsymbol{y} \mid \boldsymbol{x}) &= \sum_{t=1}^{|\boldsymbol{y}|} u(y_t \mid \boldsymbol{y}_{<t}, \boldsymbol{x}) = \sum_{t=1}^{|\boldsymbol{y}|} \left(f_t - \log Z_{t+1}\right) \notag \\
    &= \sum_{t=1}^{|\boldsymbol{y}|} \left(f_t - \log Z_t\right) + \log Z_1 - \bcancel{\log Z_{|\boldsymbol{y}|+1}} \notag \\
    &= \sum_{t=1}^{|\boldsymbol{y}|} \log p_\pi(y_t \mid \boldsymbol{y}_{<t}, \boldsymbol{x}) + \log Z_1 \notag \\
    &= \log p_\pi(\boldsymbol{y} \mid \boldsymbol{x}) + \log Z_1,
\end{align}
where $p_\pi(y_t \mid \boldsymbol{y}_{<t}, \boldsymbol{x})$ is the probability of token $y_t$ given the previous sequences $(\boldsymbol{y}_{<t}, \boldsymbol{x})$. Please note that $\log Z_1$ does not depend on the particular sequence $\boldsymbol{y}$.

\paragraph{Normalized Log-Likelihood Reward}
By combining the Plackett-Luce model in Eq.~\ref{eq:pl} with the cumulative reward in Eq.~\ref{eq:sequence_reward}, the probability for preference $\tau_n$ is given by:
\begin{equation}
\label{eq:pref_modeling}
    p(\tau_n) = \prod_{i=1}^{n} \frac{\exp \left(\log p_\pi(\boldsymbol{y}^{(i)} \mid \boldsymbol{x})\right)}{\sum_{j=i}^{n} \exp \left(\log p_\pi(\boldsymbol{y}^{(j)} \mid \boldsymbol{x}) \right)}.
\end{equation}

When modeling preferences, the term $\log Z_1$ can be eliminated due to the \textit{translation invariance} property of the softmax function. Therefore, the cumulative reward simplifies to:
\begin{equation}
    \boldsymbol{r}(\boldsymbol{y} \mid \boldsymbol{x}) = \frac{1}{|\boldsymbol{y}|} \log p_\pi(\boldsymbol{y} \mid \boldsymbol{x}).
\label{eq:sim-reward}
\end{equation}
where $1/|\boldsymbol{y}|$ is a length-normalized term to avoid bias towards longer sequences \citep{meng-2024-simpo, gu-2024-minillm}.

In other words, the reward of a language model can be formalized as the average log-likelihood, which naturally reflects the inherent preferences of the language model. \mymod{Specifically, the higher the probability the model assigns to generating a response $\boldsymbol{y}$, the greater the associated reward}~\footnote{\myadd{Existing works \citep{rafailov-2023-direct, gu-2024-minillm} also propose constructing reward functions using language models. However, the reward functions we present offer advantages in training efficiency and performance, which are discussed and compared theoretically in Appendix \ref{app:diss-reward}.}}.

\begin{figure*}[ht]
    \centering
    \vspace{-4mm}
    \includegraphics[width=\textwidth, trim=0 160 0 130, clip]{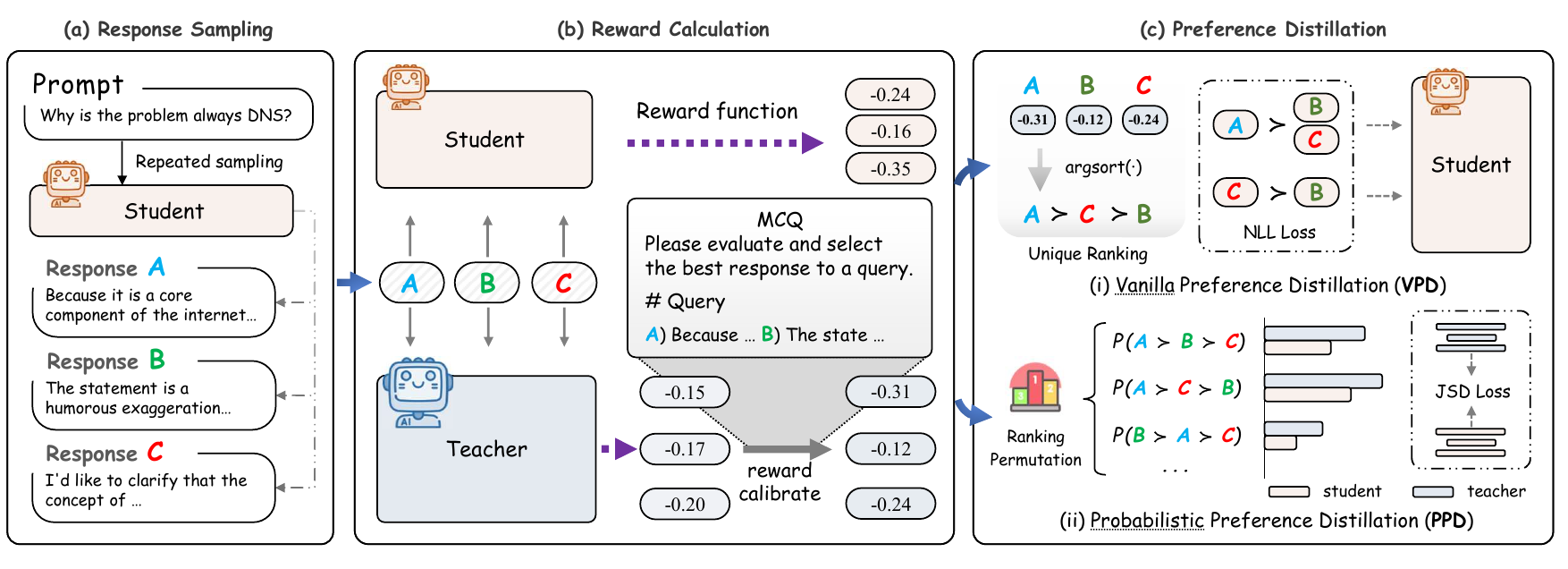}
    \vspace{-8mm}
    \caption{The overall process of the PAD contains three critical steps. The initial step involves sampling diverse responses with high temperature (§\ref{sec:phase-1}). Next, rewards for both models are computed, where the rewards of the teacher would be calibrated(§\ref{sec:phase-2}). Finally, the student is trained to mimic the teacher's preference distributions.(§\ref{sec:phase-3})}
    \vspace{-4mm}
    \label{fig:method}
\end{figure*}

\section{PAD: Preference-Aligned Distillation}

\label{sec:method}
This section outlines our PAD, which involves three key training phases (§\ref{sec:phase-1}-\ref{sec:phase-3}), followed by the introduction of a preference decomposition strategy to accelerate the training process (§\ref{sec:phase-4}).

\subsection{Diverse Response Generation}
\label{sec:phase-1}

As the first step, taking prompt $\boldsymbol{x}$ as input, we directly sample $n$ responses $Y_n$ from the student model $\pi^{\text{stu}}$ through repeated sampling. To enhance response diversity, we apply a higher decoding temperature of 0.8. This approach offers two key advantages. First, enabling the generation of higher-quality responses. Existing works have shown that as the number of repeated samples increases, the likelihood of the model generating better answers across various tasks, such as \mymod{mathematics and coding} \citep{wang-2023-selfconsistency, roz-2024-codellama, brown-2024-largelanguagemonkeysscaling}.  Second, mitigating the exposure bias. Exposure bias arises from the mismatch between training and inference, where the model is trained on ground truth contexts but relies on its own predictions during inference, leading to error accumulation. 
\mymod{Following \citet{gu-2024-minillm, agarwal2024onpolicy}, we train the student model on self-generated responses to reduce this bias.}

\subsection{Reward Calculation and Calibration}
\label{sec:phase-2}

Given a prompt $\boldsymbol{x}$ and its corresponding list of responses $Y_n$ from the previous step, we calculate the rewards for both the teacher and student models for each response $\boldsymbol{y}_i \in Y_n$ using Equation (\ref{eq:sim-reward}). These rewards, denoted as $\boldsymbol{r}^{\text{tch}}(\boldsymbol{y}_i)$ and $\boldsymbol{r}^{\text{stu}}(\boldsymbol{y}_i)$, represent the models' average log-likelihood for each response. However, language models often suffer from \textit{miscalibration}, where the assigned likelihoods do not accurately reflect the actual quality of the sequences \citep{zhao-2023-calibrating}. For instance, phrases such as "pros and cons" and "cons and pros" convey the same meaning, but the former may be more frequent in the training data, leading the model to assign it a higher probability. This miscalibration poses a challenge: if the teacher's reward is miscalibrated, aligning the student model to the teacher may propagate this issue.

To address this, we leverage insights from \citet{knowno2023} and \citet{pmlr-v239-ren23a}, who demonstrate that Multiple-Choice Question (MCQ) selection probabilities better capture response quality than sequence likelihoods. We introduce the MCQ selection probability to calibrate the teacher model's reward.\footnote{\myadd{A discussion of other calibration methods is provided in Appendix \ref{app:diss-calibrate}.}} Specifically, each response $\boldsymbol{y}_i \in Y_n$ is randomly mapped to a choice within a predefined set $C_n$ (e.g., $C_3 = \{\text{`A'}, \text{`B'}, \text{`C'}\}$), and we compute the (token-level) probability of selecting each choice:
\begin{equation}
\label{eq:select}
    p_{\text{sel}}(\boldsymbol{y}_i) = p(c_i \mid Y_n, C_n, \boldsymbol{x}),
\end{equation}
where $c_i$ corresponds to the choice associated with response $\boldsymbol{y}_i$.

We then calibrate the reward for each response by combining the normalized log-likelihood reward with the selection probability:
\begin{equation}
    \hat{\boldsymbol{r}}^{\text{tch}}(\boldsymbol{y}) = (1-\alpha) \boldsymbol{r}^{\text{tch}}(\boldsymbol{y}) + \alpha \log p_{\text{sel}}(\boldsymbol{y}),
\end{equation}
where the reward calibration ratio $\alpha \in [0,1]$ is a hyperparameter that balances the influence of the original reward and the MCQ selection probability.

\subsection{Preference Distillation}
\label{sec:phase-3}

Based on different ways of modeling teacher preferences, we employ two losses to distillation: the vanilla preference loss $\mathcal{L}_\text{VPD}$, and the probabilistic preference loss $\mathcal{L}_\text{PPD}$.

\paragraph{Vanilla Preference Distillation (VPD)}
Following \citet{rafailov-2023-direct, song-2024-pro}, the preference is modeled as a unique ranking. Specifically, we obtain ranking $\tau_n$ of the responses $Y_n$ by sorting them according to their rewards $\hat{r}^{\text{tch}}$. The student model is then trained with negative log-likelihood (NLL) loss to maximize the probability of teacher preference using Eq.~\ref{eq:pref_modeling}.
\begin{align}
    \mathcal{L}_\text{VPD} = \sum_{i=1}^n \log \frac{\exp\left( \beta \boldsymbol{r}^{\text{stu}}(\boldsymbol{y}^{(i)}) \right)}{\sum_{j=i}^n \exp\left( \beta\boldsymbol{r}^{\text{stu}}(\boldsymbol{y}^{(j)}) \right)},
\end{align}
where $\beta$ is a hyperparameter that controls the scaling of the reward difference.

\paragraph{Probabilistic Preference Distillation (PPD)}
Inspired by \citet{cao-2007-listnet}, we treat the teacher's rewards as uncertain indicators of preference, which means any preference ranking is assumed to be possible but has a different likelihood. 

The preference distribution over all possible rankings for the teacher is expressed as:
\begin{equation}
    \forall \tau_n \in \mathcal{T},\ p_{\pi^{\text{tch}}}(\tau_n) = \prod_{i=1}^n \frac{\exp\left(\beta\hat{\boldsymbol{r}}^{\text{tch}}(\boldsymbol{y}^{(i)}) \right)}{\sum_{j=i}^n \exp\left(\beta\hat{\boldsymbol{r}}^{\text{tch}}(\boldsymbol{y}^{(j)}) \right)},
\end{equation}
where $\mathcal{T}$ denotes the set of all possible rankings. The student’s preference distribution $p_{\pi^{\text{stu}}}(\tau_n)$ is modeled similarly.

We then employ the Jensen-Shannon divergence (JSD) loss to align the student's and teacher's preference distributions:
\begin{equation}
\label{eq:ppd}
    \mathcal{L}_\text{PPD}=
    \frac{1}{2} \left[\mathcal{D}_\text{KL}(\pi^{\text{tch}} || \pi^\text{mix}) + \mathcal{D}_\text{KL}(\pi^{\text{stu}} || \pi^\text{mix}) \right],    
\end{equation}
where mixed distribution $\pi^\text{mix} = (\pi^{\text{tch}} + \pi^{\text{stu}}) / 2$, and $\mathcal{D}_\text{KL}(\cdot \| \cdot)$ is the Kullback-Leibler divergence (KLD). Specifically, the KLD between the teacher's preference distribution and the mixed distribution is defined as:
\begin{equation*}
\mathcal{D}_\text{KL}(\pi^{\text{tch}} || \pi^\text{mix})= \sum_{\tau_n \in \mathcal{T}} p_{\pi^{\text{tch}}}(\tau_n) \log \frac{p_{\pi^{\text{tch}}}(\tau_n)}{p_{\pi^{\text{mix}}}(\tau_n)}.
\end{equation*}
Similarly, $\mathcal{D}_\text{KL}(\pi^{\text{stu}} || \pi^\text{mix})$ is computed in the same way. By aligning the student's preference distribution with the teacher's, the student model not only learns the correct preference ranking but also captures the teacher’s confidence in these rankings.

\subsection{Preference Decomposing Strategy}
\label{sec:phase-4}

In our PAD, the number of sampled responses, i.e., the sample size $n$, plays a pivotal role. A larger $n$ allows for a more macro comparison among responses, reduces the variance introduced by sampling, and increases the likelihood of generating high-quality responses \citep{brown-2024-largelanguagemonkeysscaling}. However, as $n$ increases, the computational cost of both sampling and forward propagation also rises. Particularly when modeling preference distributions, the complexity grows factorially, making the computation unfeasible when $n$ becomes large.

To reduce the computational cost, we propose a preference decomposition strategy. This strategy breaks down the preference of a large batch of responses into the preferences of multiple smaller batches, allowing the training process to be split into several iterative rounds, thereby reducing the overall computational load.

\paragraph{Decomposing Preference Modeling}
Given a preference ranking $\tau_n$, we define a preference decomposition function $\phi$ to decompose it into $k$ sub-preferences, such that $\phi(\tau_n) = \{\tau_m^{(1)}, \tau_m^{(2)}, \dots, \tau_m^{(k)}\}$.
Assuming that these sub-preferences are independent, we simplify the probability of the complete preference to the probability of the \textit{decomposed preferences} as follows:
\begin{equation}
\label{eq:approx}
p(\tau_n) \xrightarrow{\text{simplify}}  p(\phi(\tau_n)) = \prod_{i=1}^{k} p(\tau_m^{(i)}).
\end{equation}

\begin{figure}[!t]
    \centering
    \vspace{-2mm}
    \includegraphics[scale=0.48]{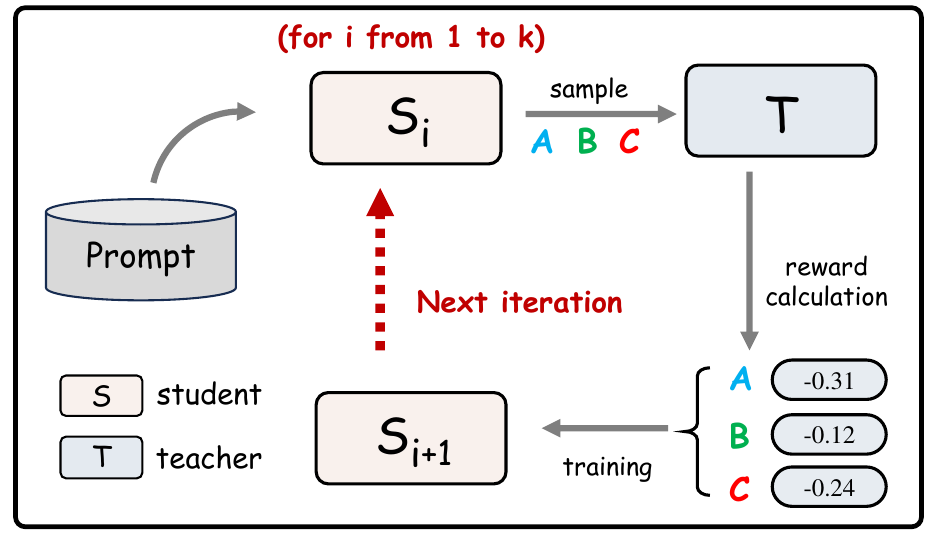}
    \vspace{-7mm}
    \caption{Iterative Distillation Process. }
    \label{fig:iter_distll}
    \vspace{-4mm}
\end{figure}

Hence, we use decomposed preferences as the learning objective, for VPD, its NLL loss for decomposed preferences as
\begin{equation}
\label{eq:log_approx}
\log p(\phi(\tau_n)) = \sum_{i=1}^{k} \log p(\tau_m^{(i)}).
\end{equation}

This shows that the distillation loss for a large batch of responses is equivalent to the sum of losses over multiple smaller batches.\footnote{A similar decomposition can be applied to Preference Propagation Decoding (PPD), with the detailed proof provided in Appendix \ref{app:theoretical}. } Based on this insight, we adopt the \textit{Iterative Distillation Process}, as depicted in Figure \ref{fig:iter_distll}. In this approach, the distillation process is divided into \(k\) iterations, each applied to smaller batches of size \(m\). This reduces the complexity of modeling the preference distribution from \(O(n!)\) to \(O(k \cdot m!)\), thereby significantly lowering the computational cost of training.

\begin{table*}[htbp]
\vspace{-2mm}

\centering
\small
\setlength{\tabcolsep}{0.95em}
\renewcommand{\arraystretch}{1.2}
\begin{tabular*}{\textwidth}{cc ll lll}
\toprule
\multirow{2}{*}{\textbf{Model Families}} & \multirow{2}{*}{\textbf{Method}} & \multicolumn{2}{c}{\textbf{Alpaca-Eval 2.0}} & \textbf{Arena-Hard} & \textbf{MT-Bench} & \textbf{GSM8K} \\ \cline{3-7} 
& & \textbf{LC (\%)} & \textbf{WR (\%)} & \textbf{WR (\%)} & \textbf{Score} (1$\sim$10) & \textbf{Acc. (\%)} \\ \hline
\multirow{12}{*}{\textsc{Gemma-2}} & Teacher (9B) & 55.27 & 42.50 & 61.16 & 6.99 & 87.41 \\
& Student (2B) & 39.51 & 41.99 & 37.55   & 6.70 & 51.63 \\\cline{2-7}
& Standard KD & 41.67 \textcolor{gray}{\scriptsize{(↑2.2})} & 45.24 \textcolor{gray}{\scriptsize{(↑3.3})} & 52.36 \textcolor{gray}{\scriptsize{(↑14.8})} & 6.78 \textcolor{gray}{\scriptsize{(↑0.1})} & 54.37 \textcolor{gray}{\scriptsize{(↑2.7})} \\
& SeqKD & 42.91 \textcolor{gray}{\scriptsize{(↑3.4})} & 46.44 \textcolor{gray}{\scriptsize{(↑4.4})} & 54.87 \textcolor{gray}{\scriptsize{(↑17.3})} & 6.88 \textcolor{gray}{\scriptsize{(↑0.2})} & 55.72 \textcolor{gray}{\scriptsize{(↑4.1})} \\
& MiniLLM & 42.97 \textcolor{gray}{\scriptsize{(↑3.5})} & 48.32 \textcolor{gray}{\scriptsize{(↑6.3})} & 55.75 \textcolor{gray}{\scriptsize{(↑18.2})} & 6.88 \textcolor{gray}{\scriptsize{(↑0.2})} & 55.26 \textcolor{gray}{\scriptsize{(↑3.6})} \\\cline{2-7}
& DPO & 43.77 \textcolor{gray}{\scriptsize{(↑4.3})} & 54.02 \textcolor{gray}{\scriptsize{(↑12.0})} & 57.43 \textcolor{gray}{\scriptsize{(↑19.9})} & 6.87 \textcolor{gray}{\scriptsize{(↑0.2})} & 57.07 \textcolor{gray}{\scriptsize{(↑5.4})} \\
& SimPO & 44.94 \textcolor{gray}{\scriptsize{(↑5.4})} & 54.16 \textcolor{gray}{\scriptsize{(↑12.2})} & 58.64 \textcolor{gray}{\scriptsize{(↑21.1})} & 6.91 \textcolor{gray}{\scriptsize{(↑0.2})} & 57.24 \textcolor{gray}{\scriptsize{(↑5.6})} \\
& PRO & 45.87 \textcolor{gray}{\scriptsize{(↑6.4})} & 56.48 \textcolor{gray}{\scriptsize{(↑14.5})} & 58.95 \textcolor{gray}{\scriptsize{(↑21.4})} & \underline{6.96} \textcolor{gray}{\scriptsize{(↑0.3})} & 58.83 \textcolor{gray}{\scriptsize{(↑7.2})} \\\cline{2-7}
& PAD w/ $\mathcal{L}_\text{VPD}$ & \underline{46.13} \textcolor{gray}{\scriptsize{(↑6.6})} & \underline{57.94} \textcolor{gray}{\scriptsize{(↑16.0})} & \underline{59.07} \textcolor{gray}{\scriptsize{(↑21.5})} & 6.93 \textcolor{gray}{\scriptsize{(↑0.2})} & \underline{59.06} \textcolor{gray}{\scriptsize{(↑7.4})} \\
& PAD w/ $\mathcal{L}_\text{PPD}$  & \textbf{49.62} \textcolor{gray}{\scriptsize{(↑10.1})} & \textbf{59.50} \textcolor{gray}{\scriptsize{(↑17.5})} & \textbf{60.00} \textcolor{gray}{\scriptsize{(↑22.4})} & \textbf{7.02} \textcolor{gray}{\scriptsize{(↑0.3})} & \textbf{59.29} \textcolor{gray}{\scriptsize{(↑7.7})} \\\hline\hline
\multirow{12}{*}{\textsc{LLaMA-3}} & Teacher (8B) & 37.01 & 38.93 & 52.66 & 7.00 & 84.00 \\
& Student (3B) & 27.82 & 29.02 & 31.70 & 6.42 & 57.09 \\\cline{2-7} 
& Standard KD & 29.11 \textcolor{gray}{\scriptsize{(↑1.3})} & 29.60 \textcolor{gray}{\scriptsize{(↑0.6})} & 41.68 \textcolor{gray}{\scriptsize{(↑10.0})} & 6.49 \textcolor{gray}{\scriptsize{(↑0.1})} & 59.15 \textcolor{gray}{\scriptsize{(↑2.1})} \\
& SeqKD & 29.48 \textcolor{gray}{\scriptsize{(↑1.7})} & 30.04 \textcolor{gray}{\scriptsize{(↑1.0})} & 42.52 \textcolor{gray}{\scriptsize{(↑10.8})} & 6.53 \textcolor{gray}{\scriptsize{(↑0.1})} & 60.94 \textcolor{gray}{\scriptsize{(↑3.8})} \\
& MiniLLM & 30.05 \textcolor{gray}{\scriptsize{(↑2.2})} & 30.38 \textcolor{gray}{\scriptsize{(↑1.4})} & 42.21 \textcolor{gray}{\scriptsize{(↑10.5})} & 6.67 \textcolor{gray}{\scriptsize{(↑0.3})} & 60.35 \textcolor{gray}{\scriptsize{(↑3.3})} \\\cline{2-7}
& DPO & 31.42 \textcolor{gray}{\scriptsize{(↑3.6})} & 32.01 \textcolor{gray}{\scriptsize{(↑3.0})} & 44.71 \textcolor{gray}{\scriptsize{(↑13.0})} & 6.62 \textcolor{gray}{\scriptsize{(↑0.2})} & \underline{61.63} \textcolor{gray}{\scriptsize{(↑4.5})} \\
& SimPO & \underline{32.74} \textcolor{gray}{\scriptsize{(↑4.9})} & \underline{32.46} \textcolor{gray}{\scriptsize{(↑3.4})} & 44.85 \textcolor{gray}{\scriptsize{(↑13.2})} & 6.73 \textcolor{gray}{\scriptsize{(↑0.3})} & 61.22 \textcolor{gray}{\scriptsize{(↑4.1})} \\
& PRO & 32.11 \textcolor{gray}{\scriptsize{(↑4.3})} & 32.23 \textcolor{gray}{\scriptsize{(↑3.2})} & 45.09 \textcolor{gray}{\scriptsize{(↑13.4})} & 6.71 \textcolor{gray}{\scriptsize{(↑0.3})} & 61.47 \textcolor{gray}{\scriptsize{(↑4.4})} \\\cline{2-7}
& PAD w/ $\mathcal{L}_\text{VPD}$ & 32.71 \textcolor{gray}{\scriptsize{(↑4.9})} & 32.34 \textcolor{gray}{\scriptsize{(↑3.3})} & \underline{45.23} \textcolor{gray}{\scriptsize{(↑13.5})} & \underline{6.77} \textcolor{gray}{\scriptsize{(↑0.3})} & 61.35 \textcolor{gray}{\scriptsize{(↑4.3})} \\
& PAD w/ $\mathcal{L}_\text{PPD}$  & \textbf{33.61} \textcolor{gray}{\scriptsize{(↑5.8})} & \textbf{32.55} \textcolor{gray}{\scriptsize{(↑3.5})} & \textbf{46.73} \textcolor{gray}{\scriptsize{(↑15.0})} & \textbf{6.84} \textcolor{gray}{\scriptsize{(↑0.4})} & \textbf{62.24} \textcolor{gray}{\scriptsize{(↑5.1})} \\
\bottomrule
\end{tabular*}
\vspace{-3mm}
\caption{Main results with the Gemma-2 and LLaMA-3 Models.}
\vspace{-5mm}
\label{tab:main}
\end{table*}

\section{Experiment}

\subsection{Setup}

\paragraph{Models}
We evaluate two model families in our main experiments: 1) \textsc{Gemma-2} Models\footnote{\url{https://ai.google.dev/gemma}} \citep{gemmateam2024improvingopen} include \textsc{Gemma-2-9B-It} as teacher and \textsc{Gemma-2-2B-It} as the student, and 2) \textsc{LLaMA-3} Models\footnote{\url{https://ai.meta.com/blog/meta-llama-3/}} \citep{dubey2024llama3herdmodels} includes \textsc{LLaMA-3.1-8B-Instruct} as teacher and \textsc{LLaMA-3.2-3B-Instruct} as student.

\paragraph{Training}
\mymod{Training data is sourced from \textsc{UltraFeedback}\footnote{\url{https://huggingface.co/datasets/argilla/ultrafeedback-binarized-preferences-cleaned}} \citep{cui2023ultrafeedback}, comprising around 60k preference samples across a diverse range of tasks, including mathematical reasoning and open-ended writing. We filter out samples that exceed the models' context length, set the number of sampled responses $n = 4$, and apply a reward calibration ratio $\alpha = 0.8$ to mitigate teacher model bias. Training is conducted for one epoch by default, with further details provided in Appendix \ref{app:train}.}

\paragraph{Evaluation}
We assess our models on four benchmarks: AlpacaEval 2.0 \citep{alpaca-eval}, MT-Bench \citep{zheng-2023-judging}, Arena-Hard \citep{arenahard-2024}, and GSM8K \citep{cobbe-2021-math}, which evaluate a model's conversational versatility across a range of tasks. For AlpacaEval, we report both the raw win rate (WR) and the length-controlled win rate (LC), the latter being robust to verbosity. For Arena-Hard, we report the win rate (WR)\footnote{Please note that for AlpacaEval 2.0 and Arena-Hard, we employ \textsc{LLaMA-3.1-70B-Instruct} as the judge model, which achieved capabilities comparable to GPT-4 Turbo on the judge test of AlpacaEval while being more cost-effective and faster.}. For MT-Bench, we present the average MT-Bench score, evaluated by GPT-4 Turbo. Detailed evaluation settings can be found in Appendix \ref{app:eval}.

\paragraph{Baselines}
We compare PAD with two types of baselines: 
1) Traditional Knowledge Distillation, which aims to learn the teacher's distribution at the logits level, including \textbf{Standard KD} \citep{hinton2015distill}, \textbf{SeqKD}
 \citep{kim-rush-2016-sequence}, and \textbf{MiniLLM} \citep{gu-2024-minillm}; 
2) Preference Knowledge Distillation, which aims to transfer the teacher's preference knowledge to the student model. Under the ``Teacher-as-Annotator'' paradigm, we choose \textbf{DPO} \citep{tunstall-2024-zephyr}, \textbf{SimPO} \citep{meng-2024-simpo}, and \textbf{PRO} \citep{song-2024-pro} as baselines. A detailed description of these baselines can be found in Appendix \ref{app:baseline}.

\subsection{Main Result}
\label{sec:main_result}
\mymod{Table \ref{tab:main} summarizes the experimental results across several benchmarks.}The primary finding is that the student models trained with PAD consistently outperform both their initial counterparts and existing approaches. Specifically, PAD achieves over a 20\% improvement on Alpaca-Eval 2.0 and Arena-Hard, highlighting its ability to more effectively capture the teacher's preferences, thereby aligning more closely with human values.

Preference distillation methods offer significant advantages over traditional KD methods. Traditional KD approaches, such as Standard KD and SeqKD, yield modest improvements (1-3\%) in Alpaca-Eval 2.0 LC scores. In contrast, preference distillation methods like DPO and SimPO show more substantial gains, particularly in aligning with human preferences, as evidenced in Alpaca-Eval 2.0 and Arena-Hard benchmarks. These findings align with the work of \citet{tunstall-2024-zephyr}.

\mymod{Notably, PAD with $\mathcal{L}_\text{PPD}$ outperforms PAD with $\mathcal{L}_\text{VPD}$ across multiple benchmarks, including Alpaca-Eval 2.0 LC (49.62 vs. 46.13) and MT-Bench (7.02 vs. 6.93). For the \textsc{Gemma-2} model family, the student trained with $\mathcal{L}_\text{PPD}$ even surpasses its teacher in MT-Bench, scoring 7.02 compared to the teacher’s 6.99. The key advantage of PPD over VPD lies in its preference modeling approach: by capturing the full preference distribution rather than just a simple ranking, PPD provides more nuanced supervisory signals, thereby better reflecting subtle human preferences—an aspect often overlooked by existing distillation methods.}

\subsection{Analysis}

We analyze the impact of our proposed Preference Decomposing Strategy and Reward Calibration. To assess the generalization capability of PAD, we further investigate its performance when the teacher and student models belong to different families. A more detailed analysis is provided in Appendix \ref{app:additional}.

\paragraph{Effect of Preference Decomposing Strategy}

\begin{table}[!tp]
\small
\centering
\setlength{\tabcolsep}{1.4em}
\renewcommand{\arraystretch}{1.2}
\begin{tabular}{c ll}
\hline
\multirow{2}{*}{\bf Strategy} & \textbf{Alpaca-Eval 2} & {\textbf{GPU Hours}} \\ \cline{2-3} 
& LC (\%) & A800 \\ \hline
\multicolumn{3}{c}{\textit{Total Sample Size: 4}} \\
1 $\times$ 4   & 49.42 & 12.32  \\ 
2 $\times$ 2   & 48.94 \textcolor{gray}{\scriptsize{(↓0.5)}} & 12.35 \textcolor{gray}{\scriptsize{(↑0.0\%)}}  \\ 
\hline
\multicolumn{3}{c}{\textit{Total Sample Size: 8}} \\
1 $\times$ 8   & 50.57 & 31.51  \\
2 $\times$ 4   & 51.42 \textcolor{gray}{\scriptsize{(↑0.8)}} & 27.98 \textcolor{gray}{\scriptsize{(↓12\%)}}  \\
\hline

\multicolumn{3}{c}{\textit{Total Sample Size: 12}} \\
1 $\times$ 12 & 51.55 & 52.74  \\ 
3 $\times$ 4 & 52.67 \textcolor{gray}{\scriptsize{(↑1.1)}} & 42.07 \textcolor{gray}{\scriptsize{(↓20\%)}}  \\
\hline
\end{tabular}
\vspace{-2mm}
\caption{\myadd{Decomposition Strategy Improves Performance While Reducing Cost. ``Strategy'' such as ``1 × 4'' indicates that the number of training iterations is 1, and the sample size per iteration is 4.}}
\vspace{-5mm}
\label{tab:iter}
\end{table}

\begin{figure}[!tbp]
    \vspace{-2mm}
    \includegraphics[scale=0.43]{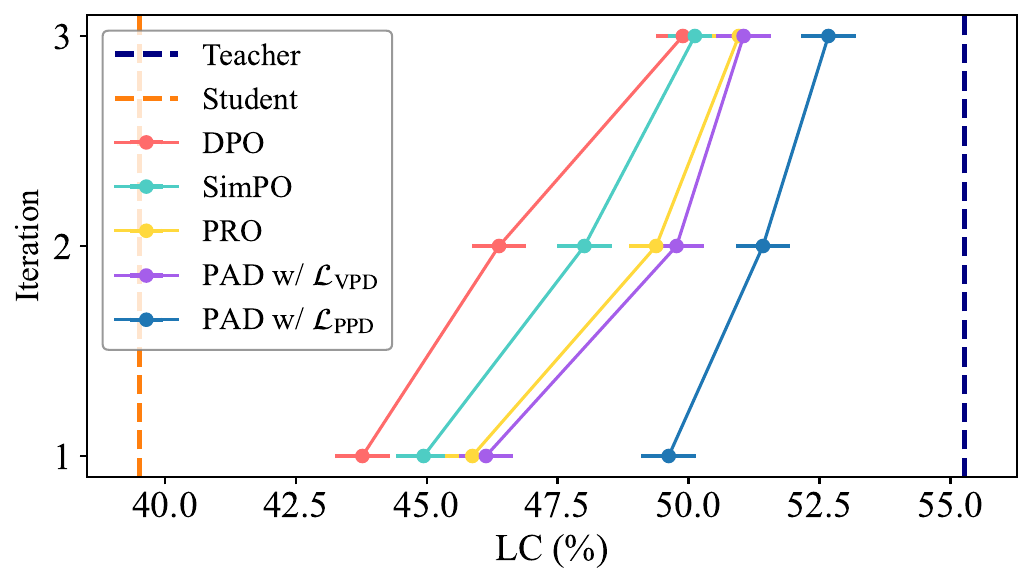}
    \vspace{-4mm}
    \caption{\mymod{Alpaca-Eval LC with different iterations.}}
    \label{fig:lc_iter}
    \vspace{-2mm}
\end{figure}

\mymod{We investigated the impact of the iterative distillation process on performance and training time using the preference decomposing strategy. Table \ref{tab:iter} presents the effects of varying iteration counts and sample sizes. For a sample size of 4, decomposing the sampling process into two iterative steps does not reduce training time, as the low complexity of modeling the distribution with fewer samples renders the effect negligible. However, when the sample size increases to 12, adopting a three-iteration decomposition reduces training time by 20\%, demonstrating that the preference decomposing strategy becomes more advantageous as the sample size grows, thereby accelerating training. Notably, decomposing the sampling into multiple iterations does not result in a significant performance drop, indicating that the strategy effectively balances efficiency with stable performance.}

\mymod{Additionally, we examined the iterative distillation process as a continuous learning method. As shown in Figure \ref{fig:lc_iter}, we compared three high-performing baseline methods—DPO, SimPO, and PRO. The results indicate that the iterative distillation process consistently improves performance across all methods, with PAD achieving the best results, highlighting its superior effectiveness.}

\begin{table}[!tp]
\small
\centering
\setlength{\tabcolsep}{.8em}
\renewcommand{\arraystretch}{1.4}

\begin{tabular}{ccc}
\hline
\textbf{Method} & \textbf{Calibration} & \textbf{Alpaca-Eval LC (\%)} \\ 
\hline
Student & - & 39.51 \\
\hline
PAD w/ $\mathcal{L}_\text{VPD}$       & No                         & 40.36 \textcolor{gray}{\scriptsize{(↑0.8)}}                          \\
PAD w/ $\mathcal{L}_\text{VPD}$       & Yes                        & 46.13 \textcolor{gray}{\scriptsize{(↑6.6)}} \\
\hline
PAD w/ $\mathcal{L}_\text{PPD}$       & No                         & 41.59 \textcolor{gray}{\scriptsize{(↑2.0)}}
                          \\
PAD w/ $\mathcal{L}_\text{PPD}$       & Yes                        &\ 49.62 \textcolor{gray}{\scriptsize{(↑10.1)}}                         \\
\bottomrule
\end{tabular}
\vspace{-0.15cm}
\caption{\myadd{Ablation study on the Reward Calibration}}
\vspace{-0.4cm}
\label{tab:reward_ablation}
\end{table}

\begin{figure}[!tbp]
    \centering
    \vspace{-1mm}
    \includegraphics[scale=0.5]{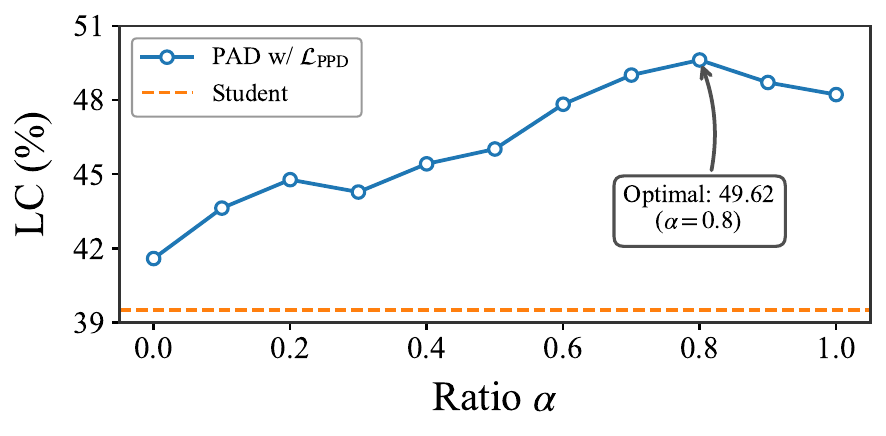}
    \vspace{-4mm}
    \caption{\mymod{Alpaca-Eval LC Win Rate with different $\alpha$}}
    \vspace{-1mm}
    \label{fig:ablation}
\end{figure}

\paragraph{Influence of Reward Calibration}
\myadd{We conduct ablation experiments on \textsc{Gemma-2-2B-It} with Reward Calibration, as summarized in Table \ref{tab:reward_ablation}. The results show a significant performance improvement, with an 8.03 LC gain on PAD with $\mathcal{L}_\text{PPD}$.}
\mymod{To further assess the impact of reward calibration, we vary the calibration ratio $\alpha$ and examine its effect on performance, as depicted in Figure \ref{fig:ablation}. Increasing $\alpha$ leads to consistent performance gains, peaking at $\alpha = 0.8$. Beyond this point, further increases in $\alpha$ result in diminishing returns. For applications with lower performance demands, setting $\alpha = 1$ offers a practical trade-off by avoiding the computational cost of calculating the teacher's log-likelihood.}

\paragraph{Heterogeneous Study} 

Previous experiments typically involve teacher and student models from the same family, sharing a common vocabulary and similar architectures. To evaluate the generalization ability of our PAD method, we extend these experiments by using teacher and student models from different families. As shown in Table \ref{tab:heterogeneous}, our approach consistently outperforms existing methods, demonstrating robust generalization. 
\myadd{Furthermore, a comparison with results in Table \ref{tab:main} reveals that using a stronger teacher model, \textsc{Gemma-9B}, significantly improves the performance of the student model (\textsc{LLaMA-3B}). Specifically, with \textsc{Gemma-9B} as the teacher, the student achieves an LC of 51.96\%, compared to only 33.61\% with \textsc{LLaMA-8B}. This suggests a positive correlation between the teacher's performance and the student's post-distillation performance, aligning with findings from \citet{gu-2024-minillm, lee-2024-rlaif}.}

\begin{table}[!tp]
\small
\centering
\setlength{\tabcolsep}{1.3em}
\renewcommand{\arraystretch}{1.4}
\begin{tabular}{cll}
\hline
\multirow{2}{*}{\textbf{Method}} & \multicolumn{2}{c}{\textbf{Alpaca-Eval LC (\%)}} \\ \cline{2-3} 
 & \textsc{LLaMA} & \textsc{Qwen} \\ \hline
Student & 27.82 & 9.37  \\ \hline
DPO   & 49.42 \textcolor{gray}{\scriptsize{(↑21.6)}} & 16.66 \textcolor{gray}{\scriptsize{(↑7.3)}}    \\ 
SimPO & 49.78 \textcolor{gray}{\scriptsize{(↑22.0)}} & 18.08 \textcolor{gray}{\scriptsize{(↑8.7)}}   \\ 
PRO   & 50.18 \textcolor{gray}{\scriptsize{(↑22.4)}} & 18.57 \textcolor{gray}{\scriptsize{(↑9.2)}}   \\ \cline{1-3} 
PAD w/ $\mathcal{L}_\text{VPD}$ & 50.45 \textcolor{gray}{\scriptsize{(↑22.6)}} & 18.83 \textcolor{gray}{\scriptsize{(↑9.4)}} \\
PAD w/ $\mathcal{L}_\text{PPD}$ & \textbf{51.96} \textcolor{gray}{\scriptsize{(↑24.1)}} & \textbf{20.57} \textcolor{gray}{\scriptsize{(↑11.2)}}  \\ \hline
\end{tabular}
\vspace{-2mm}
\caption{\myadd{Heterogeneous Distillation Study. We use \textsc{Gemma-2-9B-It} as the teacher and \textsc{LLaMA-3.2-3B-Instruct}/\textsc{Qwen-2.5-1.5B-Instruct} as the student.}}
\vspace{-5mm}
\label{tab:heterogeneous}
\end{table}

\section{Related Work}

\paragraph{Traditional Knowledge Distillation}
Knowledge distillation (KD), introduced by \citet{hinton2015distill}, primarily aims at model compression by training a smaller student model to mimic the output behavior of a larger teacher model \citep{kim-rush-2016-sequence, liang-2021-mixkd, zhang-2023-blindly, gu-2024-minillm, agarwal2024onpolicy}.
\citet{kim-rush-2016-sequence} extended KD to machine translation by training students on sequences generated by teachers in order to imitate teacher behavior. More recently, \citet{gu-2024-minillm} advanced KD using reverse KL divergence on student -generated sequence to mitigate exposure bias, improving student model performance. A key feature of these methods is that distillation is performed over the shared vocabulary of both teacher and student models. Our PAD eliminates this limitation, enabling effective distillation with different vocabularies.

\paragraph{Preference Knowledge Distillation}
Motivated by the observation that large models have achieved a high degree of alignment with human values and preferences, many efforts focus on distilling preference knowledge from large models to smaller ones \citep{bai-2022-constitutional, cui2023ultrafeedback, lee-2024-rlaif, yuan-2024-selfreward, tunstall-2024-zephyr, yang2024rlcd}. \citet{bai-2022-constitutional} first introduced this concept, also known as Reinforcement Learning from AI Feedback (RLAIF), where teacher models annotate response pairs from the student to create a preference dataset for training a reward model. \citet{tunstall-2024-zephyr} further utilized teacher-annotated preferences with Direct Preference Optimization (DPO) \citep{rafailov-2023-direct}, streamlining the training of student models.  These approaches follow the "Teacher-as-Annotator" paradigm. The annotated preference datasets generated through this paradigm can be directly employed with methods such as DPO, SimPO \citep{meng-2024-simpo}, and PRO \citep{song-2024-pro}, enabling preference optimization of student models. However, a significant limitation of these methods lies in their reliance on unique ranking, which constrains their ability to model nuanced preferences. In contrast, our PAD treats modeling preference knowledge as a distribution over all possible preferences, enabling nuanced alignment for the student and teacher models.

\section{Conclusion}

In this paper, we introduced the Preference-Aligned Distillation (PAD) framework, which models the teacher's preference knowledge as a probability distribution over all potential preferences. This supervisory signal enables the student model to capture subtle distinctions between responses. \mymod{Experimental results on the \textsc{Gemma-2} and \textsc{LLaMA-3} model families show that PAD outperforms both traditional knowledge distillation and existing preference distillation methods across four benchmark tasks, highlighting its capacity for learning in-depth human preferences.}

\section*{Limitations}
Our research has several limitations. Firstly, the generalization capability is insufficient as we have not conducted experiments on larger-scale teacher and student models, primarily due to limited computational resources. Secondly, sampling multiple responses consumes more computational overhead. However, because SLMs have relatively smaller parameter sizes, this overhead remains comparatively modest.
Thirdly, our method requires token-level probabilities, which are unavailable in some black-box models.

\bibliography{custom}

\appendix

\newpage

\clearpage
\section{Decomposing Probabilistic Preference Distillation}
\label{app:theoretical}
Substituting Eq. \ref{eq:approx} into the KLD:
\begin{align*}
     & \sum_{\tau_n} \left( \prod_{i=1}^{k} p_{\pi^{\text{tch}}}(\tau_m^{(i)}) \right) \log \left( \frac{\prod_{i=1}^{k} p_{\pi^{\text{tch}}}(\tau_m^{(i)})}{\prod_{i=1}^{k} p_{\pi^{\text{mix}}}(\tau_m^{(i)})} \right) \\
     =& \sum_{\tau_n} \left( \prod_{i=1}^{k} p_{\pi^{\text{tch}}}(\tau_m^{(i)}) \right) \left( \sum_{i=1}^{k} \log \frac{p_{\pi^{\text{tch}}}(\tau_m^{(i)})}{p_{\pi^{\text{mix}}}(\tau_m^{(i)})} \right)\\
     \intertext{interchange summations,}
     =& \sum_{i=1}^{k} \sum_{\tau_n} \left( \prod_{j=1}^{k} p_{\pi^{\text{tch}}}(\tau_m^{(j)}) \right) \log \frac{p_{\pi^{\text{tch}}}(\tau_m^{(i)})}{p_{\pi^{\text{mix}}}(\tau_m^{(i)})}
\end{align*}
notice that for a fixed $i$, the logarithm term only depends on $\tau_m^{(i)}$, and the product can be separated,
\begin{small}
\begin{align*}
     = \sum_{i=1}^{k} \left( \sum_{\tau_m^{(i)}} p_{\pi^{\text{tch}}}(\tau_m^{(i)}) \log \frac{p_{\pi^{\text{tch}}}(\tau_m^{(i)})}{p_{\pi^{\text{mix}}}(\tau_m^{(i)})} \right) \prod_{j \neq i} \left( \sum_{\tau_m^{(j)}} p_{\pi^{\text{tch}}}(\tau_m^{(j)}) \right)
\end{align*}
\end{small}
\vspace{-6mm}
\begin{flalign*}
     \intertext{based on the independence assumption of sub-preferences, $\sum_{\tau_m^{(j)}} p_{\pi^{\text{tch}}}(\tau_m^{(j)}) = 1$ for each $j$, }
    & = \sum_{i=1}^{k} \sum_{\tau_m^{(i)}} p_{\pi^{\text{tch}}}(\tau_m^{(i)}) \log \frac{p_{\pi^{\text{tch}}}(\tau_m^{(i)})}{p_{\pi^{\text{mix}}}(\tau_m^{(i)})}&
\end{flalign*}

Therefore, the KLD can be decomposed as:
\begin{align*}
   &\mathcal{D}_{\text{KL}}(p_{\pi^{\text{tch}}}(\tau_n) \| p_{\pi^{\text{mix}}}(\tau_n)) =\\ & \quad \quad \quad \quad \sum_{i=1}^{k} \mathcal{D}_{\text{KL}}(p_{\pi^{\text{tch}}}(\tau_m^{(i)}) \| p_{\pi^{\text{mix}}}(\tau_m^{(i)}))
\end{align*}

The JSD Loss (Eq. \ref{eq:ppd}) used in PPD is the average of two KLDs in different directions, making JSD also decomposable.

\section{Implementation Details}
\label{app:inplement}

\subsection{Training}
\label{app:train}

\begin{table}[!tp]
\small
\centering
\setlength{\tabcolsep}{1.3em}
\renewcommand{\arraystretch}{1.3}
\begin{tabular}{ccc}
\hline
\textbf{hyperparameter} & \textbf{Value} & \textbf{Searching Space} \\ \hline
$\beta$ & 10 & $[1, 2, 5, 8, 10]$  \\ 
batch size & 128 &  $[32, 64, 128]$  \\ 
warmup ratio & 0.1 &  $[0.05, 0.1]$  \\ 
\hline
\end{tabular}
\caption{The hyperparameter values in PAD training.}
\label{tab:hyper}
\end{table}
We individually search the learning rates for different model families in the range of $ [3e-7, 5e-7, 8e-7, 1e-6, 1e-5]$. As a result, the learning rate for \textsc{Gemma-2} Models is $8e-7$ and for \textsc{Llama-3} Models is $1e-6$. Table \ref{tab:hyper} shows other hyperparameters for training.
All the training experiments in this paper were conducted on 2×A800 GPUs based on the TRL repo\footnote{\url{https://github.com/huggingface/trl/tree/main}}.

\subsection{Evaluation}
\label{app:eval}
\paragraph{Data Statistics}
AlpacaEval 2 consists of 805 questions from five datasets, MT-Bench includes 80 questions across eight categories, and the recently released Arena-Hard is an enhanced version of MT-Bench, comprising 500 challenging questions. Since the training data, ultrafeedback, includes some mathematical reasoning problems, we additionally incorporate the GSM8K test set, which contains approximately 1,300 questions, to evaluate the model's mathematical abilities.

\paragraph{Judge Models} For AlpacaEval 2.0 and Arena-Hard, we employ \textsc{LLaMA-3.1-70B-Instruct} as the judge model. For MT-Bench, we employ GPT-4 Turbo as the judge model. For GSM8K, we report accuracy on the test set. Table \ref{tab:judge} presents the evaluation capability test\footnote{\url{https://github.com/tatsu-lab/alpaca_eval/tree/main/src/alpaca_eval/evaluators_configs}} of these judge models on AlpacaEval. We can see that \textsc{LLaMA-3.1-70B-Instruct} has evaluation capabilities comparable to GPT4-Turbo.

\begin{table}[!tp]
\small
\centering
\setlength{\tabcolsep}{0.6em}
\renewcommand{\arraystretch}{1.3}
\begin{tabular}{cc}
\hline
\textbf{Model} & \textbf{Human Agreement} \\ \hline
GPT4 & 69.17  \\ 
\textsc{LLaMA-3.1-70B-Instruct} & 69.10  \\ 
GPT4-Turbo & 68.09 \\
\textsc{LLaMA-3-70B-Instruct} & 67.53  \\ 
\textsc{Qwen2.5-72B-Instruct} & 67.51 \\\hline
Humans & 65.66 \\
\hline
\end{tabular}
\caption{Leaderboard of judge models in AlpacaEval.}
\label{tab:judge}
\end{table}

\subsection{Baselines}
\label{app:baseline}
For traditional knowledge distillation, we consider three baselines: 1) \textbf{Standard KD} \citep{hinton2015distill}: Fine-tunes the student model using the teacher model's logits distribution as a supervision signal, applied to golden responses.
2) \textbf{SeqKD} \citep{kim-rush-2016-sequence}: Directly fine-tunes the student model with cross-entropy loss using responses generated by the teacher model.
3) \textbf{MiniLLM} \citep{gu-2024-minillm}: Employs the teacher model's logits distribution as supervision signal while fine-tuning the student model on its own generated responses.

For preference knowledge distillation, under the "Teacher-as-Annotator" paradigm, we employ three offline preference optimization methods as baselines:
1) \textbf{DPO} \citep{rafailov-2023-direct, tunstall-2024-zephyr}: Treats the student model as a reward model, fine-tuning it based on a reward function derived from a reference model.
2) \textbf{SimPO} \citep{meng-2024-simpo}: Operates similarly to DPO but uses average log-likelihood as the optimization objective.
3) \textbf{RPO} \citep{song-2024-pro}: Extends the above approaches by optimizing with listwise preference information. For a fair comparison, we also use responses sampled from the student model for these baselines. We use the MCQ selection probability introduced in Section \ref{sec:phase-2} as the score to rank the responses. 
For the pairwise preference optimization methods DPO and SimPO, we select the responses with the maximum and minimum rewards to form preference pairs. For the listwise preference optimization method PRO, we directly sort the scores to form the preference ranking. Please kindly note that constructing preference pairs using the maximum and minimum scores of responses is a common practice \citep{cui2023ultrafeedback, meng-2024-simpo}. Moreover, our preliminary experiments indicate that splitting the entire listwise response data into multiple pairwise data and training with DPO/SimPO does not yield significant performance improvements. 

\section{Additional Experiments and Analyses}
\label{app:additional}

\paragraph{Scaling Up} 
\begin{table}[ht]
\small
\centering
\setlength{\tabcolsep}{1.3em}
\renewcommand{\arraystretch}{1.2}
\begin{tabular}{cccccc}
\hline
\multirow{2}{*}{\textbf{Method}} & \textbf{Alpaca-Eval} & \textbf{Arena-Hard} \\ \cline{2-3} 
& LC (\%) & WR (\%) \\ \hline
Teacher  & 56.89 & 76.09    \\ 
Student  & 39.51 & 37.55  \\ 
DPO   & 51.93 & 65.42  \\ 
SimPO   & 52.36 & 66.36  \\ 
PRO   & 52.45 & 68.01  \\ 
\hline
PAD w/ $\mathcal{L}_\text{VPD}$   & 53.32 & 67.38 \\
PAD w/ $\mathcal{L}_\text{PPD}$   & 55.96 & 69.90  \\ \hline
\end{tabular}
\caption{Scaling-up Study.}
\label{tab:scaling_up}
\end{table}

We use \textsc{Gemma-2-27B-It} as the teacher and \textsc{Gemma-2-2B-It} as the student. The overall performance is shown in Table ~\ref{tab:scaling_up}. When employing larger-scale teacher models, our PAD consistently and significantly enhances the ability of small models to align with human preferences. Compared to the main result (§\ref{sec:main_result}), we observe that when using a more capable teacher, the student model achieves greater performance improvements, indicating that the performance gap between the teacher and student is a key factor in determining the extent of the student's enhancement.

\begin{figure}[!tbp]
    \includegraphics[scale=0.23]{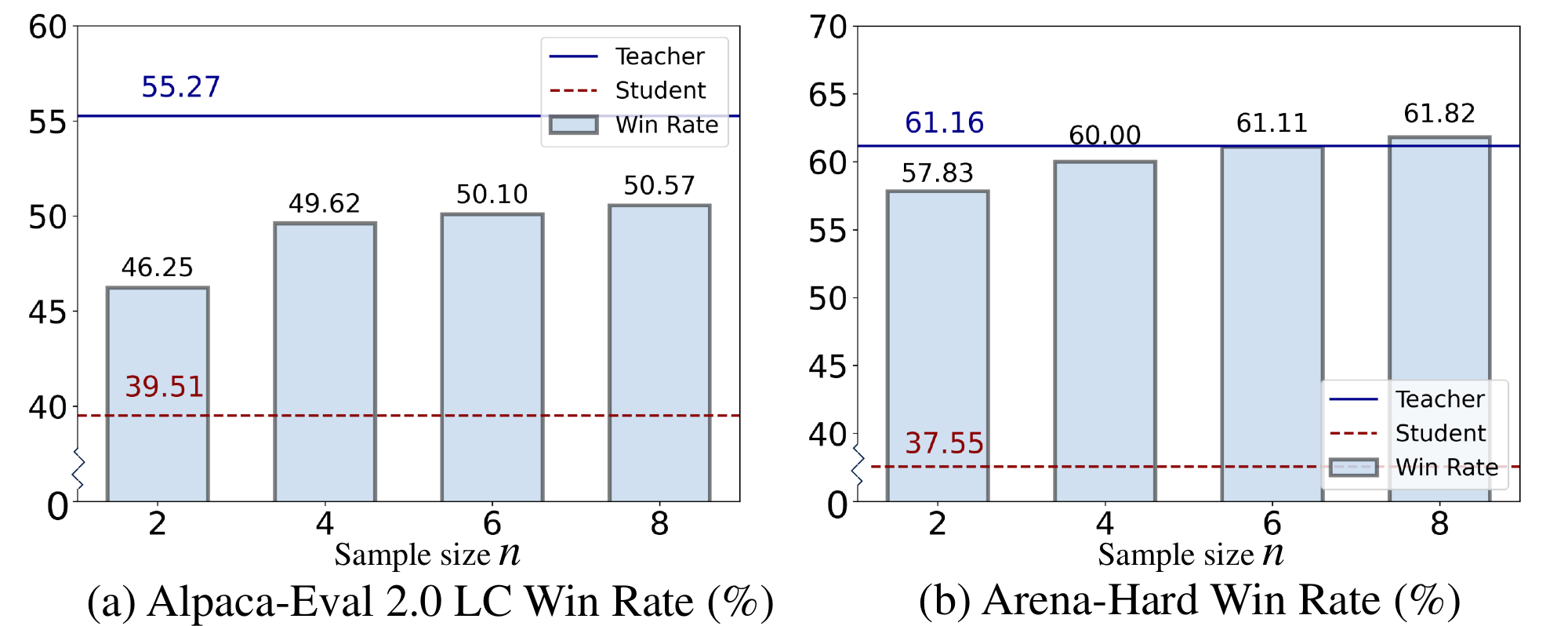}
    \caption{Win Rate with different sample size $n$.}
    \label{fig:sample}
\end{figure}

\paragraph{Effect of Sample Size} We investigate the impact of the number of sampled responses on PPD, and the results can be seen in Figure \ref{fig:sample}. We observe that as the number of sample size $n$ increases, the performance of the student model improves accordingly. This indicates that obtaining more feedback knowledge through extensive sampling from the text generation space facilitates better alignment of the student model with the teacher's preferences.

\section{Comparative Analysis of Reward Functions}
\label{app:diss-reward}

\begin{table*}[ht]
\centering
\small
\begin{tabular}{@{} l l c c c @{}}
\toprule
& \textbf{Formulation} 
& \textbf{Preference Expression} 
& \textbf{Reference-free} 
& \textbf{Aligned with Generation} 
\\
\midrule

\textbf{DPO} 
& $\mathbf{r}(\boldsymbol{y}) = \log \dfrac{p_\text{current}(\boldsymbol{y})}{p_\text{reference}(\boldsymbol{y})}$ 
& $\checkmark$ 
& $\times$ 
& $\times$ 
\\[0.8ex]

\textbf{MiniLLM} 
& $\mathbf{r}(\boldsymbol{y}) = \log \dfrac{p_\text{teacher}(\boldsymbol{y})}{p_\text{student}(\boldsymbol{y})}$ 
& $\times$ 
& $\times$ 
& $\times$ 
\\[0.8ex]

\textbf{PAD (Ours)} 
& $\mathbf{r}(\boldsymbol{y}) = \dfrac{1}{| y |} \log p_\text{current}(\boldsymbol{y})$ 
& $\checkmark$ 
& $\checkmark$ 
& $\checkmark$ 
\\[0.8ex] %

\bottomrule
\end{tabular}
\caption{Theoretical Comparison of Reward Functions}
\label{tab:reward_comparison}

\vspace{-0.5em}
\end{table*}

\myadd{For the preference distillation, existing works DPO\citep{rafailov-2023-direct} and MiniLLM\citep{gu-2024-minillm} both propose using language models to construct reward functions. 
However, there are significant differences between these methods and our PAD, particularly in the formulation of the reward function, the theoretical foundation, and how the reward is used during training. Below, we elaborate on three key distinctions.}

\subsection{Formulation of Reward Functions}

\paragraph{DPO (Direct Preference Optimization)} \myadd{constructs the reward function using the log ratio of the likelihoods between the "current model" and the "reference model" for the same response:
$$\mathbf{r}(\boldsymbol{y}) = \log \frac{p_\text{current}(\boldsymbol{y})}{p_\text{reference}(\boldsymbol{y})},$$
where $p_\text{current}$ represents the likelihood of response $\boldsymbol{y}$ under the model aligned with human preferences, and $p_\text{reference}$ typically corresponds to the likelihood from a supervised fine-tuning (SFT) model. A higher reward indicates that the current model is more likely to generate $\boldsymbol{y}$ than the reference model.}

\paragraph{MiniLLM} \myadd{is similar in spirit to DPO but uses a “teacher model” and “student model” in a distillation framework. Its reward function is based on the likelihood ratio:
$$\mathbf{r}(\boldsymbol{y}) = \log \frac{p_\text{teacher}(\boldsymbol{y})}{p_\text{student}(\boldsymbol{y})}.$$
Here, a larger reward implies that the teacher model is more inclined to generate the sequence than the student model.}
    
\paragraph{PAD (Ours)} \myadd{defines the reward function as the model’s log-likelihood of the response, averaged by the sequence length for consistency across varying lengths:
$$\mathbf{r}(\boldsymbol{y}) = \frac{1}{|y|} \log p_\text{current}(\boldsymbol{y}).$$
When the model assigns a higher probability to a response, that response receives a higher reward.}

\subsection{Theoretical Foundations and Training Processes}

\paragraph{Different Theoretical Foundations.}
\myadd{DPO is grounded in the Reinforcement Learning from Human Feedback (RLHF) framework, while MiniLLM derives its reward function from the reverse KL-divergence of the policy gradient. In contrast, PAD is based on Inverse Reinforcement Learning (IRL), where we demonstrate that the average log-likelihood can function as a reward, effectively capturing the model’s intrinsic preferences.}

\paragraph{Roles in the Loss Function.}
\myadd{DPO aims to directly maximize the margin between "good" and "bad" responses within its loss function. MiniLLM, on the other hand, treats the reward function as a scaling factor for the SFT loss, where responses with higher rewards experience a more significant increase in likelihood. PAD utilizes the reward function to derive a preference distribution over responses, enabling a student model to emulate the teacher model's preferences. Notably, unlike DPO and PAD, MiniLLM's reward function does not explicitly encode preference.}

\paragraph{Reference Model Requirement.}
\myadd{Both DPO and MiniLLM necessitate the simultaneous use of two models (current/reference or teacher/student) for reward computation. In contrast, PAD requires only the log-likelihood of the current model, eliminating the need for a reference model. This simplification not only streamlines the theoretical framework but also reduces computational cost in practice.}

\paragraph{Alignment with the Generation Stage.}
\myadd{The reward functions in DPO and MiniLLM generally do not align with the probability distribution used during the final text-generation phase. In contrast, PAD’s reward function is directly aligned with the model’s log-likelihood, ensuring consistency between the training and inference.}

\subsection{Empirical Evaluation}

\myadd{Beyond the theoretical distinctions outlined in Table \ref{tab:reward_comparison}, the choice of reward significantly influences performance across various benchmarks. As illustrated in the main results of Table \ref{tab:main}, PAD, which employs average log-likelihood as its reward function, demonstrates superior performance gains.}

\section{Discussion of Reward Calibration}
\label{app:diss-calibrate}

\subsection{Existing Methods}  
\myadd{
Recent studies have shown that large language models (LLMs) are well-calibrated for multiple-choice question-answering and true/false evaluation tasks \cite{kadavath2022languagemodelsmostlyknow, openai-2024-gpt4technicalreport, robinson2023leveraging}, suggesting that these models exhibit better calibration on token-level scores. In addition to the MCQ-based calibration method, we also compare it with the \textit{P(True)} method proposed by \citet{kadavath2022languagemodelsmostlyknow}.
}
\paragraph{P(True)} \myadd{involves asking the model whether a candidate answer is correct. If the answer is correct, the model outputs \texttt{True}, otherwise \texttt{False}, with the probability of \texttt{True} representing the likelihood of the response being correct. Formally, for an input $x$ and a response $y$, this probability is given by:}
$$
p_{\text{true}}(x, y) = p(\text{Yes} \mid x, y).
$$

\myadd{
A variant, \textbf{P(True) with reference}, incorporates all candidate responses:}
$$
p_{\text{true}}(x, y, Y) = p(\text{Yes} \mid x, y, Y).
$$
\myadd{
Notably, the P(True) method is more computationally intensive than the MCQ-based method, as it requires multiple queries to determine the probability values for different responses, whereas MCQ-based calibration can map responses to options and obtain probabilities in a single query.}

\subsection{Experimental Comparison}  
\myadd{We conducted experiments on the \textbf{Alpaca-eval Evaluator} test set \citep{alpaca-eval} to assess the alignment of various calibrated reward functions with human preferences. This test set, consisting of approximately 2.5k data points, is commonly used to evaluate LLM performance as judges, where each data point includes a pair of responses ranked by human preferences. We applied different reward calibration methods to rank these preference pairs and evaluated their alignment with human preferences.}

\myadd{The evaluation metrics include \textit{Human Agreement} and \textit{Prob. prefer longer}.}

\paragraph{Human Agreement:} \myadd{This metric measures the alignment between human and model rankings of response pairs. Given $n$ response pairs ($y_g$ for the good response and $y_b$ for the bad response), the score is calculated as:}
\begin{equation*}
\begin{split}
\text{score} = 1 - \frac{1}{n} \sum_{i=1}^{n} \biggl[ 
    &\left| \text{rank}_i^\text{human}(y_g) - \text{rank}_i^\text{pred}(y_g) \right| \\
    + &\left| \text{rank}_i^\text{human}(y_b) - \text{rank}_i^\text{pred}(y_b) \right| \biggr]
\end{split}
\end{equation*}
\myadd{with a score closer to 1 indicating stronger alignment with human preferences.}

\paragraph{Prob. prefer longer:} \myadd{This metric measures the likelihood that the model's preferred response is longer than the alternative, indicating a potential length bias in the model's preferences.}

\myadd{We evaluated the performance of two models, Gemma-2-9B-It and Llama-3.1-8B-Instruct, with the results presented in Table \ref{tab:calibrate_cmp}. These results show that the MCQ-based calibration achieves the highest alignment with human preferences while exhibiting a relatively lower bias toward longer responses.}

\begin{table}[ht]
\small
\centering
\setlength{\tabcolsep}{1em}
\renewcommand{\arraystretch}{1.2}
\begin{tabular}{lccccc}
\hline
\multirow{2}{*}{\textbf{Method}} & \textbf{Agreement} & \textbf{Prob. prefer longer} \\ \cline{2-3} 
& (\%) & (\%) \\ \hline
\multicolumn{3}{c}{\textsc{Gemma-2}} \\
MCQ & 68.21 & 65 \\
P(True) & 61.27 & 52 \\
P(True) w/ ref. & 67.75 & 76 \\
\hline
\multicolumn{3}{c}{\textsc{Llama-3}} \\
MCQ & 65.59 & 75 \\
P(True) & 63.12 & 60 \\
P(True) w/ ref. & 65.47 & 76 \\
\hline
\end{tabular}
\caption{Alpaca Evaluator Performance Comparison.}
\label{tab:calibrate_cmp}
\end{table}

\myadd{We further examined the impact of different reward calibration methods on PAD with PPD Loss using two benchmarks, Alpaca-Eval 2.0 and Arena-Hard. The results, shown in Table \ref{tab:calibrate_cmp_eval}, demonstrate that the MCQ-based method consistently outperforms other methods, achieving the highest performance on both benchmarks.
}

\begin{table}[ht]
\centering
\small
\centering
\setlength{\tabcolsep}{1.2em}
\renewcommand{\arraystretch}{1.2}
\begin{tabular}{lcc}
\hline
\multirow{2}{*}{\textbf{Method}} & \textbf{Alpaca-Eval} & \textbf{Arena-Hard} \\ \cline{2-3} 
& LC (\%) & WR (\%) \\ 
\hline
MCQ & \textbf{49.62} & \textbf{59.50} \\
P(True) & 44.09 & 53.96 \\
P(True) w/ ref. & 48.59 & 57.74 \\
\hline
\end{tabular}
\caption{Impact of Different Reward Calibration Methods on PAD with Gemma-2}
\label{tab:calibrate_cmp_eval}
\end{table}

\myadd{These findings highlight the significant impact of the chosen reward calibration method on PAD performance. The MCQ-based method not only aligns more closely with human preferences but also improves the effectiveness of PAD training, leading to better alignment in the distilled model.}

\end{document}